\RequirePackage{fix-cm}
\documentclass[smallextended]{svjour3}       
\smartqed  

\usepackage{graphicx}
\usepackage{subfigure}
\usepackage{amsmath}
\usepackage{booktabs}
\usepackage{longtable}
\usepackage{setspace}
\usepackage{epstopdf}
\usepackage{multirow}
\usepackage{changepage}
\usepackage{threeparttable}
\usepackage[misc]{ifsym}
\usepackage{amsfonts,amssymb}
\usepackage{amsfonts}
\usepackage{multirow}
\usepackage{color}
\usepackage{algorithm,algpseudocode}
\usepackage{comment}
\usepackage{url}
\begin{document}

\title{Scalable Multi-view Clustering with Graph Filtering
}


\author{
        Liang Liu 
        \textsuperscript{1} 
        \and Peng Chen \textsuperscript{2} 
         \and Guangchun Luo \textsuperscript{3} 
        \and Zhao Kang \textsuperscript{1}
        \and Yonggang Luo \textsuperscript{4}
        \and Sanchu Han
        \textsuperscript{4}
}

\institute{
\Letter Zhao Kang\\
\email{Zkang@uestc.edu.cn}\\
 \at
 {1} School of Computer Science and Engineering, University of Electronic Science and Technology of China, Chengdu, China.
 \at
 {2} Jangsu Automation Research Institute, Lianyungang, China.
 \at
 {3} School of Information and Software Engineering, University of
Electronic Science and Technology of China, Chengdu, China.
 \at
 {4} Chongqing Changan Automobile Co., Ltd, Chongqing, China.
 \\
}



\date{Received: date / Accepted: date}

\maketitle

\begin{abstract}
With the explosive growth of multi-source data, multi-view clustering has attracted great attention in recent years. Most existing multi-view methods operate in raw feature space and heavily depend on the quality of original feature representation. Moreover, they are often designed for feature data and ignore the rich topology structure information. Accordingly, in this paper, we propose a generic framework to cluster both 
attribute and graph data with heterogeneous features. It is capable of exploring the interplay between feature and structure. Specifically, we first adopt graph filtering technique to eliminate high-frequency noise to achieve a clustering-friendly smooth representation. To handle the scalability challenge, we develop a novel sampling strategy to improve the quality of anchors. Extensive experiments on attribute and graph benchmarks demonstrate the superiority of our approach with respect to state-of-the-art approaches. 
\end{abstract}

\section{Introduction}
With the rapid advances in information technology, many data are collected from various views or appear in multiple modalities, which form the so-called multi-view data. In other words, the same object can be described from different angles with heterogeneous features \cite{kang2021structured}. For instance, an image can be represented by different types of features, such as Gabor, HOG, GIST, and LBP; a text news can be translated into multiple languages. Each individual view contains some specific property that should be explored \cite{chao2021survey,hou2018safe}. Nowadays, many data are also described in graph or network, which is a popular data structure to characterize interdependent systems \cite{yu2021learning}. For example, an academic network can represent the relations among authors. In reality, multiple types of relations exist, e.g., co-author and co-paper relations in academic network. Hence, multilayer graph or multiplex network (i.e., multiview graph) are often applied to describe such systems and each layer accounts for one type of relation. In practice, nodes in the graph are often attached with attributes and this kind of data are named attributed graph. Therefore, it is desirable to take full advantage of available information for better performance in the downstream tasks.  

To analyze those kinds of multiview data, clustering is a widely used technique to unveil meaningful patterns of samples or nodes by dividing them into disjoint groups \cite{huang2021robust,zhu2019PR}. To address above heterogeneous challenge, many researchers have been dedicating efforts in developing methods that are able to effectively discover a common cluster pattern shared by various views \cite{mi2022multi,liu2021refined}. A straightforward way is to concatenate all multiple views features and apply conventional single-view clustering methods upon it. This naive approach, however, completely ignore the correlation among multiple views \cite{chen2020multi}. Therefore, more complex methods are proposed, which are supposed to explore the consensus and complementary information across multiple views. For instance, \cite{kumar2011cotraining} proposes a co-regularization technique to minimize the disagreement between each pair of views. Inspired by co-training strategy, \cite{kumar2011co} finds the clusterings that agree across the views. Roughly speaking, the existing methods can be divided into two categories. The first class of methods try to fuse the features \cite{wang2020parallel,liu2021incomplete}, while the others integrate clusterings \cite{kang2020partition}. Nevertheless, these methods are susceptible to poor quality data, which lead to degraded clustering performance. 

Instead of operating on the raw features, some recent approaches manage to learning in latent space. For example, \cite{zhang2017latent} assumes that multiple views are originated from one underlying latent representation and reconstructs multi-view data to obtain a common subspace representation, upon with subspace clustering technique is implemented. Motivated by the success of deep neural networks, some deep multi-view clustering techniques are also developed. For instance, \cite{li2019deep} utilizes deep autoencoder to learn latent representations shared by multiple views and 
employs adversarial training to disentangle the latent space. To deal with graph data, \cite{fan2020one2multi} leverages a graph autoencoder to learn node embeddings of one selected view and reconstruct multiple graphs. Hence, its performance heavily depends on the chosen view and it fails to make full use of available data. By contrast, \cite{cheng2020multi} uses multiple graph autoencoders to extract multiple embeddings and find a common clustering. Although these methods have attractive performance, they have high computational complexity. Thus, the clustering of multi-view graph data is still at a nascent stage.  

From above analysis, we can observe that the clustering methods for feature and graph data are developed individually and there is no general framework that is suitable for both types of data. Therefore, they cannot exploit the rich feature and topology information in attributed graph. Moreover, they perform poor on noisy data and have high complexity. To this end, in this paper, we propose a novel and generic clustering method for various multi-view data: Scalable Multi-view Clustering with graph filtering (SMC). Compared to existing works, it has two distinct properties. First, it achieves a clustering-friendly representation for each view. According to clustering assumption, the feature values of sample points belonging to the same class are similar. This can be realized by graph filtering technique developed in signal processing community, which renders the signal smooth. Second, to reduce the computation complexity, a novel sampling strategy is introduced based on the importance of nodes. Comprehensive experiments and analysis demonstrate the superiority of our method.

\section{Related Work}
\subsection{Notations}
Without loss of generality, we define multi-view data $\mathcal{G}= (\mathcal{V},\xi_1,\cdots,\xi_v,X^1,\cdots,X^v)$, where $\mathcal{V}$ denotes the set of $n$ nodes, $\xi_v$ is the edge set, $X^v=\{\emph{\textbf{x}}_1^v,\cdots,\emph{\textbf{x}}_n^v\}^\top\in\mathbb{R}^{n\times d_v}$ is the $v$-th feature matrix of nodes. $\xi^v$ can be combined into the topology structure of graph $A^v$, and $a_{ij}^v$ reflects the relationship between node $i$ and node $j$.
If a dataset don't have a graph associated with it, we can build one for each view as discussed later. Then,  $D^v=\emph{diag}(d_1,\cdots,d_n)\in\mathbb{R}^{n\times{n}}$ denotes the degree matrix  of $A^v$, where $d_{i} = \sum_{j=1}^{n}a_{ij}^v$. The symmetrically normalized Laplacian can be derived as $L^v = I - (D^v)^{-\frac{1}{2}}(A^v)(D^v)^{-\frac{1}{2}}$.
\subsection{Multi-view Clustering}
Among various multi-view clustering methods, graph-based approaches often produce more impressive performance. AMGL \cite{Nie2016Parameter} is a multi-view spectral clustering model with an auto-weighting mechanism. MLRSSC \cite{Brbi2017Multi} learns a joint subspace representation across all views with low-rank and sparsity constraints. RMSC \cite{xia2014robust} pursues a latent low-rank transition probability matrix and obtains clustering results by standard Markov chain method. PwMC and SwMC \cite{nie2017self} learn a shared graph from input graphs by applying a novel self-weighting strategy. MSC\_IAS \cite{wang2019multi} applies Hilbert–Schmidt Independence Criterion (HSIC) to maximize the learned similarity with its corresponding intact space. LMVSC \cite{kang2019large} is proposed to tackle the scalability challenge of multi-view subspace clustering. Specifically, it learns a smaller similarity matrix by the idea of anchor. Based on it, SMVSC \cite{chen2021smoothed} is further proposed to enjoy a smooth data representation. To process multi-view graph, PMNE \cite{liu2017principled} is developed to learn embeddings. MNE \cite{zhang2018scalable} is a scalable multi-view network
embedding method. Nevertheless, these methods are designed for either feature or graph data and are not applicable to attributed graph data.  

Latter, O2MAC \cite{fan2020one2multi} is developed to cluster multi-view attributed graph. Similarly, HAN \cite{HAN} presents a
graph neural network for heterogeneous graph by combining with attention mechanism. However, they often fail to fully exploit the rich semantic information of multi-view data. Recently, MvAGC \cite{lin2021graph} is proposed for multi-view attributed graph clustering and shows impressive performance with a shallow approach. It is flexible to explore high-order relations among nodes \cite{lin2021multi}. In this paper, we aim to bring a generic framework for both feature and graph data.

\section{Methodology}
\subsection{Graph Filtering}
Given a feature matrix $X=[\emph{\textbf{x}}_1,\cdots,\emph{\textbf{x}}_n]^\top\in\mathbb{R}^{n\times d}$ with $n$ samples and $d$ features, it can be treated as $d$ $n$-dimensional graph signals. A natural signal should be smooth on nearby nodes in term of the underlying graph, i.e., nearby nodes have close feature values \cite{dong2019learning}. From another perspective, a smooth signal will contain more low-frequency basis signals than high-frequency ones. In general, high-frequency components are regarded as noise. Hence, a smooth signal is supposed to be free of noise and thus benefits downstream analysis. To recover a clean signal $\bar{X}$, we can solve the following problem:

\begin{equation}
    	\min_{\bar{X}}\|\bar{X} - X\|_F^2 + \mu Tr(\bar{X}^{\top}L\bar{X}),
    \label{optimization}
\end{equation}
where $\mu \textgreater 0$ is a balance parameter. The first term is a fidelity term and the second term is graph Laplacian regularization, which means that $\bar{x}_i$ and $\bar{x}_j$ should be close if samples $i$ and $j$ are similar in original space. Taking the first-order derivative of the objective function w.r.t. $\bar{X}$ and setting it to
zero, we have
\begin{equation}
    	\bar{X} = (I+\mu L)^{-1}X.
    \label{derivative}
\end{equation}
Above solution involves matrix inversion whose time complexity is $O (n^3)$. Therefore, we approximate $\bar{X}$ by its first-order Taylor series expansion, namely, $\bar{X} = (I-\mu L)X$. More generally, filtering with $k$ times can be written as:
\begin{equation}
    	\bar{X} = (I-\mu L)^{k}X.
    \label{filtering}
\end{equation}
When $\mu=1/2$, it goes back to the previously used filter $\bar{X} = (I-L/2)^{k}X$. Therefore, Eq.(\ref{filtering}) is a generalization of previous filter \cite{lin2021graph}. Because it has a tunable parameter $\mu$, it gives us more flexibility on real datasets \cite{pan2021multi}.
For multi-view data, we could obtain a smooth representation for each view, i.e., $\bar{X^i}=(I - \mu{L^{i}})^k{X^i}$. $k$-order graph filtering captures the long-distance structure information of graph by aggregating features of neighbors up to $k$th-order, which makes adjacent nodes have similar feature values. In other words, graph filtering encodes the structure information into feature. Therefore, it provides an elegant way to integrate the rich feature and structure information.

\subsection{Clustering}
It is known that representation is crucial to the performance of machine learning algorithms. For clustering task, it is assumed that similar samples are more likely assigned into the same group. Therefore, it is reasonable to smooth the data before feeding them into clustering model. It has been shown that graph filtering indeed increases the distance between clusters, which in turn facilitates subsequent clustering \cite{ma2020towards}. Therefore, $\bar{X}$ can be applied in various clustering methods. In this paper, we choose subspace clustering due to its intriguing performance \cite{lv2021pseudo}. Following \cite{kang2019large,chen2021smoothed}, we utilize the  self-expressiveness property of data, i.e., each sample can be expressed as a linear combination of other samples, to learn a similarity matrix, which characterizes the similarities between any two instances \cite{zhang2020twin}. As for multi-view data $\bar{X^i}$, we obtain the similarity matrix for each view. The problem can be modeled as:
\begin{equation}
	\min_{\{Z^i\}_{i=1}^{v}} \sum_{i=1}^{v} \|{\bar{X^i}}^{\top}-{{B^i}}(Z^i)^{\top}\|_F^2+\alpha \|Z^i\|_F^2,
	\label{multi-view-model}
\end{equation}
    where $\alpha$ is a trade-off parameter and $B^i=[b_1^i,\cdots,b_m^i]\in{R^{d_i \times {m}}}$ is the anchor matrix for $i$-th view. Note that we replace $\bar{X^i}$ with ${B^i}$ in the first term to reduce the computation complexity. Previously, it constructs a $n\times n$ similarity matrix and applies spectral clustering to achieve clustering result, whose time complexity is up to $\mathcal{O}(n^3)$. In Eq. (\ref{multi-view-model}), we just learn a smaller $n\times m$ similarity matrix $Z^i$ to alleviate the computation burden. Specifically, we select $m$ representative points for each view, which are supposed to reconstruct the corresponding $\bar{X}^i$ based on $Z^i$ that characterizes the similarities between $n$ original nodes and $m$ anchors. More details about anchor selection strategy are provided in section 3.4. Eq. (\ref{multi-view-model}) admits a closed-form solution.

Next, we concatenate $Z^i$ as $\bar{Z}=[Z^1,\cdots,Z^i,\cdots,Z^v]\in \mathcal{R}^{n\times mv}$. It has been shown that the spectral embedding matrix $Q\in\mathcal{R}^{n\times g}$, which consists of $g$ eigenvectors associated with the largest $g$ eigenvalues of $\sum_i {Z^iZ^{i^\top}}/v$, can be achieved by applying singular value decomposition (SVD) on $\bar{Z}$ \cite{kang2019large}. Finally, K-means is implemented on $Q$ to obtain the final $g$ partitions. 


 \subsection{Anchor Selection}
    It is easy to see from (\ref{multi-view-model}) that the choice of anchors will impact the solution. The mainstream approach adopts K-means or random sampling, which makes sense for general feature data. In particular, we run K-means on $\bar{X^i}$ and let $g$ cluster centers form $B^i$. However, this approach could be suboptimal for graph data since each node has different importance and K-means treats all nodes equally. It is natural to sample anchors based on the importance of nodes. Let's define $q:\mathcal{V}\rightarrow{R^+}$ as the importance measure function and assign the probability to node $i$ that it will be chosen as the first member of anchor set $\mathcal{M}$:
    \begin{equation}
        \label{sampling}
        p_i=\frac{q(i)^\gamma}{\sum_{j\in{\mathcal{V}}}(q(j)^\gamma)},
    \end{equation}
    where $\gamma \in R^+$, which makes the distribution sharp (for $\gamma \textgreater 1$) or smooth (for $\gamma \textless 1$). After that, we select $m-1$ distinct nodes without replacement. Specifically, each left node $i\in{\mathcal{V}\setminus{\mathcal{M}}}$ is chosen with a probability 
    $p_i/{\sum_{j\notin{\mathcal{M}}}}p_j$ as the second anchor, and so on until  $\lvert \mathcal{M} \rvert = m$. For simplicity, the total degree of each node, i.e., $q(i)=\sum_{k=1}^v\sum_{j\in{\mathcal{V}}}A^k_{ij}$, is employed to characterize its importance.  

\begin{algorithm}[H]
\caption{SMC}\label{algorithm}
		\textbf{Input}: multi-view feature $\{X^i\}_{i\in[1,v]}$, or multi-view graph $\{A^i\}_{i\in[1,v]}$ with feature $\{X^i\}_{i\in[1,v]}$\\
		\textbf{Parameter}: filter order $k$, filter parameter $\mu$, trade-off parameter $\alpha$, anchor number {$m$}, \\ cluster number $g$, sampling parameter $\gamma$ \\
		\textbf{Output}: $g$ partitions
        \begin{algorithmic}[1]
        \If{only multi-view feature is available}
                \State Build a graph for each view based on certain graph construction method
                \State Apply $k$ times graph filter on $\{X^i\}_{i\in[1,v]}$ to obtain the representation $\bar{X}^1, \cdots, \bar{X}^i, \cdots, \bar{X}^v$
                \State Sample $m$ anchors by applying K-means on $ \bar{X}^i$ to construct $B^i$.
        \Else
                \State Apply $k$ times graph filter on $\{X^i\}_{i\in[1,v]}$ to obtain the representation $\bar{X}^1, \cdots, \bar{X}^i, \cdots, \bar{X}^v$
                \State Sample $m$ anchors with indexes \emph{ind} and then choose $m$ rows from $\bar{X}^i$ to build $B^i$.
        \EndIf

        \State Calculate ${Z^i}\in \mathbb{R}^{n \times m}$ in Eq. (\ref{multi-view-model}), which is composed of ${\bar{Z}} \in\mathcal{R}^{n\times mv}$   
        \State Compute $Q$ by performing SVD on ${\bar{Z}}$
        \State Apply K-means to Q
        \end{algorithmic}
\end{algorithm}
\subsection{Time Complexity}
    The overall procedures for our proposed method is outlined in Algorithm \ref{algorithm}. Suppose $N$ is the number of nonzero elements of adjacency matrix $A$, the time complexity of performing graph filtering is $\mathcal{O}(Ndkv)$, where $d=\sum_i d_i$. Sampling anchors process takes $\mathcal{O}(mv)$ and solving $Z^i$ takes $\mathcal{O}(nm^3v)$. In addition, the computation of $Q$ costs $\mathcal{O}(m^3v^3+2mvn)$ and the subsequent K-means consumes $\mathcal{O}(ng^2)$. In real applications, $A$ is often sparse, thus $N \ll n^2 $. Furthermore, $m,v \ll n$. Thus, the overall complexity could be linear to the sample number $n$. It is worth mentioning that our algorithm is iteration-free and makes it appealing in practice. The implementation of our method is public available at: \url{https://github.com/EricliuLiang/SMC}. 
\section{Experiment}
\subsection{Datasets}
We perform extensive experiments on several benchmark datasets to demonstrate the effectiveness of our model. We choose widely used multi-view datasets, including Handwritten, Caltech-7, Caltech-20, and Citeseer. Handwritten\footnote{https://archive.ics.uci.edu/ml/datasets/Multiple+Features} and Caltech\footnote{http://www.vision.caltech.edu/ImageDatasets/Caltech101/} are image datasets of digits and objects respectively. They both have six different views. Citeseer\footnote{https://lig-membres.imag.fr/grimal/data.html} is a citation network whose nodes denote publications with two views. To apply graph filtering, we employ probabilistic neighbor method \cite{nie2016constrained} to obtain a graph for these datasets. Furthermore, some attributed graph data are also evaluated, including ACM, DBLP and IMDB \cite{fan2020one2multi}. They are heterogeneous graph that have different types of relationship between the same set
of nodes. Specifically, ACM and DBLP are paper networks that contain two types (co-paper and co-subject) and three types (co-author, co-conf, and co-term) of relationship respectively. IMDB is a movie network with two views. The detailed statistics of the datasets are shown in Table \ref{dataset}.

\begin{table}[!hbtp]
	\begin{center}
		\caption{The information of datasets used in the experiments. The value in parenthesis denotes feature dimension or the number of edges. }
		\label{dataset} 
		\resizebox{1.0\columnwidth}{!}{
			\begin{tabular}{llll}
				\hline
				
				{View} &{Handwritten} & {Caltech-7/Caltech-20}  & {Citeseer}  \\\hline
				1& Profile Correlations (216) & Gabor(48) &  Citation Links (3312) \\
				2& Fourier Coefficients (76) & Wavelet moments (40)&  Words Presence (3703)\\
				3& Karhunen Coefficients (64)  &CENTRIST (254)& -\\
				4 &  Morphological (6) &HOG (1984)& -\\
				5& Pixel Averages (240)  & GIST (512)& - \\
				6& Zernike Moments (47)  &LBP (928)& -  \\\hline
				Data samples & 2000 & 1474/2386& 3312 \\
				Cluster number& 10 & 7/20 & 6 \\
				\hline
				
				{View} &{ACM} & {DBLP}  & {IMDB}  \\ \hline 
				1 & co-paper (29,281)  & co-author (11,113) &  co-actor (98,010)  \\
				2& co-subject (2,210,761) & co-conf (5,000,495) &  co-director (21,018)\\
				3& -  & co-term (6,776,335) & -\\ \hline 
            
				Node & 3025 (1830) & 4057 (334)& 4780 (1232) \\
				Cluster number & 3 & 4 & 3 \\
				\hline
		\end{tabular}}
	\end{center}
\end{table}

\subsection{Experimental Setup}
To  have  a  convincing  comparison  between  our method  and  existing  approaches, we select some representative methods. For multi-view datasets, we compare with AMGL \cite{Nie2016Parameter}, MLRSSC \cite{Brbi2017Multi},
MSC\_IAS \cite{wang2019multi}, and recently proposed LMVSC \cite{kang2019large}, SMVSC \cite{chen2021smoothed}. For multi-view graph datasets, we compare both multi-view and single-view methods. LINE \cite{tang2015line} and GAE \cite{Kipf2016VariationalGA} are typical single-view methods, while MNE \cite{zhang2018scalable}, PMNE \cite{liu2017principled}, RMSC \cite{xia2014robust}, PwMC and SwMC \cite{nie2017self}, O2MAC \cite{fan2020one2multi}, HAN \cite{HAN}, and MvAGC \cite{lin2021graph} are targeted for multi-view graph data.

Clustering performance is evaluated by five commonly used metrics, including accuracy (ACC), normalized  mutual  information (NMI), purity (PUR), F1-score(F1), Adjusted Rand Index (ARI).

\subsection{Multi-view Feature Data Result}
The results on the multi-view feature data are given in Table \ref{ImagaResult}. For most measures, our proposed SMC achieves the best performance. Compared with SMVSC, the result is also improved. This is attributed to the introduction of $\mu$ and our filter is adaptive to different data. With respect to other methods that don't employ graph filtering, the improvement is more significant, which verifies the advantage of smooth representation. 
\begin{table*}[htbp]\scriptsize
		\caption{Clustering performance on multi-view feature data. }
		\label{ImagaResult}
        \centering
		\setlength{\tabcolsep}{6mm}{
			\begin{tabular}{c| c| c| c| c}
				\hline
				{Datasets}&{Method}  & ACC & NMI & PUR  \\
				\hline
				\multirow{5}{*}{Handwritten}&AMGL \cite{Nie2016Parameter} & 84.60& 87.32& 87.10\\
				&MLRSSC \cite{Brbi2017Multi}  & 78.90& 74.22& 83.75 \\
				&MSC\_IAS \cite{wang2019multi}  & 79.75& 77.32& 87.55\\
				&LMVSC \cite{kang2019large} & 91.65& 84.43& 91.65\\
				&SMVSC \cite{chen2021smoothed} & 94.30 &	88.95 &	94.30\\
				&SMC ($k=1$)&\textbf{96.20}&	\textbf{91.76}&	\textbf{96.20}\\				
				\hline
				\multirow{5}{*}{Caltech-7}&AMGL \cite{Nie2016Parameter}
				& 45.18& 42.43& 46.74\\
				&MLRSSC \cite{Brbi2017Multi} & 37.31& 21.11& 41.45\\ 
				&MSC\_IAS \cite{wang2019multi} & 39.76& 24.55& 44.44\\
				&LMVSC \cite{kang2019large} & 72.66& 51.93& 75.17\\
				&SMVSC \cite{chen2021smoothed} & 73.54 &	\textbf{52.04} &	84.87\\\
				&SMC ($k=1$)&\textbf{78.69}& 48.29 &	\textbf{88.60}\\		
				\hline
				\multirow{5}{*}{Caltech-20}&AMGL \cite{Nie2016Parameter}
				& 30.13& 40.54& 31.64\\ 
				&MLRSSC \cite{Brbi2017Multi} & 28.21& 26.70& 30.39\\ 
				&MSC\_IAS \cite{wang2019multi} & 31.27& 31.38& 33.74\\
				&LMVSC \cite{kang2019large} & 53.06& 52.71 & 58.47\\
				&SMVSC \cite{chen2021smoothed} & 56.92 &	51.90&	\textbf{64.42}\\
				&SMC ($k=1$)&\textbf{57.16}&	\textbf{54.58}&	62.66\\	
				\hline
				\multirow{5}{*}{Citeseer}&AMGL \cite{Nie2016Parameter} & 16.87& 0.23& 16.87\\ 
				&MLRSSC \cite{Brbi2017Multi} & 25.09& 02.67& 63.70\\ 
				&MSC\_IAS \cite{wang2019multi} & 34.11& 11.53& \textbf{80.76}\\
				&LMVSC \cite{kang2019large} & 52.26& 25.71& 54.46\\
				&SMVSC \cite{chen2021smoothed} & 55.40&	25.57&	57.27\\
				&SMC ($k=1$)& \textbf{56.00}&	\textbf{29.85}&	56.00\\	
				\hline
			\end{tabular}
		}
	\end{table*}

\subsection{Heterogeneous Graph Data Result}
For the graph data scenario, due to the adoption of Eq. (\ref{sampling}), we report the mean value after twenty runs in Table \ref{GraphResult}. We can observe that our model produces competitive and attractive results. In particular, our method shows advantage over deep neural networks-based techniques, e.g., O2MAC and HAN. 
\begin{table}[h]
    \centering
    \caption{Clustering results on multi-view graph data. The '-' means that the method runs out of memory. }
    \label{GraphResult}
    \renewcommand\arraystretch{1.2}
    \scalebox{0.55}{
\begin{tabular}{c|c|c|c|c|c|c|c|c|c|c|c|c}
\hline
\multirow{2}{*}{\textbf{Method}} & \multicolumn{4}{c|}{ACM}          & \multicolumn{4}{c|}{DBLP}         & \multicolumn{4}{c}{IMDB}         \\ \cline{2-13} 
                        & ACC    & F1     & NMI    & ARI    & ACC    & F1  & NMI    & ARI    & ACC    & F1     & NMI    & ARI    \\ \hline
 LINE \cite{tang2015line} & 0.6479 & 0.6594 & 0.3941 & 0.3433 & 0.8689 & 0.8546 & 0.6676 & 0.6988 & 0.4268 & 0.287 & 0.0031 & -0.009 \\ \hline
 GAE \cite{Kipf2016VariationalGA} & 0.8216 & 0.8225 & 0.4914 & 0.5444 & 0.8859 & 0.8743 & 0.6925 & 0.741 & 0.4298 & 0.4062 & 0.0402 & 0.0473 \\ \hline
 MNE \cite{zhang2018scalable} & 0.637 & 0.6479 & 0.2999 & 0.2486 & - & - & - & - & 0.3958 & 0.3316 & 0.0017 & 0.0008 \\ \hline
PMNE(n) \cite{liu2017principled} & 0.6936 & 0.6955 & 0.4648 & 0.4302 & 0.7925 & 0.7966 & 0.5914 & 0.5265 & 0.4958 & 0.3906 & 0.0359 & 0.0366 \\ \hline
PMNE(r) \cite{liu2017principled} & 0.6492 & 0.6618 & 0.4063 & 0.3453 & 0.3835 & 0.3688 & 0.0872 & 0.0689 & 0.4697 & 0.3183 &  0.0014 & 0.0115 \\ \hline
PMNE(c) \cite{liu2017principled} & 0.6998 & 0.7003 & 0.4775 & 0.4431 & - & - & - & - & 0.4719 & 0.3882 & 0.0285 & 0.0284 \\ \hline
RMSC \cite{xia2014robust} & 0.6315 & 0.5746 & 0.3973 & 0.3312 & 0.8994 & 0.8248 & 0.7111 & 0.7647 & 0.2702 & 0.3775 & 0.0054 & 0.0018 \\ \hline
PwMC \cite{nie2017self} & 0.4162 & 0.3783 & 0.0332 & 0.0395 & 0.3253 & 0.2808 & 0.019 & 0.0159 & 0.2453 & 0.3164 & 0.0023 & 0.0017 \\ \hline
SwMC \cite{nie2017self} & 0.3831 & 0.4709 & 0.0838 & 0.018 & 0.6538 & 0.5602 & 0.376 & 0.38 & 0.2671 & 0.3714 &  0.0056 & 0.0004   \\ \hline
O2MAC \cite{fan2020one2multi} & \textbf{0.9042} & \textbf{0.9053} & \textbf{0.6923} & \textbf{0.7394} &0.9074 & 0.9013 & 0.7287 & 0.778 & 0.4502 & \textbf{0.4159} & \textbf{0.0421} & 0.0564 \\ \hline
 HAN \cite{HAN} & 0.8823 & 0.8844 & 0.5881 & 0.5933 & 0.9114 & 0.9078 & 0.7859 & 0.8124 & 0.5547 & 0.4152 & 0.0986 & 0.0856 \\
 \hline
 \hline
MvAGC \cite{lin2021graph}                  & 0.8975 & 0.8986 & 0.6735 & {0.7212}  & 0.9277 &{0.9225}  & 0.7727 &0.8276 & 0.5633 & 0.3783 & 0.0371 & 0.0940 \\ \hline

SMC ($k=1$)                  & 0.8849  & 0.8856        & 0.6360        & 0.6897       & 0.9337 & 0.9297 & 0.7822   & 0.8386   & 0.5464    & 0.4091 & 0.0397 & 0.0863  \\ \hline

SMC ($k=2$)                  & 0.8863  & 0.8869        & 0.6397        & 0.6929       & \textbf{0.9347} & \textbf{0.9304} & \textbf{0.7862}   & \textbf{0.8421}   & \textbf{0.5686}    & 0.4048 & 0.0356 & \textbf{0.0998}  \\ \hline

\end{tabular}}
\end{table}
Though the recent method MvAGC explores high-order relations, our SMC still achieves comparable and even better performance.  improvement is brought by the adaptive filter, which suits to different data. In summary, we can draw the following conclusions. First, multi-view methods outperform single-view methods due to the exploitation of rich complementary information. Second, graph filtering approaches that employ both structure and attribute information generally perform better than many others that only use one type of information. This also verifies the significance of fully 
exploiting the interactions between feature and structure. Third, importance sampling is effective, which incorporates different roles of nodes in graph. Therefore, our method obtains impressive performance for both feature and graph data.

\subsection{Time Comparison}
We also test the time consumed by different methods. The experiments are conducted on the same machine with an Intel(R) Core(TM) i7-6800k 3.40GHZ CPU, an GeForce GTX 1080 Ti GPU and 32GB RAM.
From Table \ref{multi-view-Time}, it can be seen that our method is very efficient with respect to others. Though our complexity is linear to $n$, it is also influenced by the number of anchors, which cause fluctuations on  different datasets. For heterogeneous graph data, our method is several orders of magnitude faster than many others, especially deep neural networks-based approaches, as shown in Table \ref{Time}. This makes our method attractive in practice. Even compared with MvAGC, our method also runs a little bit faster.
\begin{table}[htbp] \small
		\caption{Time comparison on feature datasets.}
		\label{multi-view-Time}
		\centering
		\setlength{\tabcolsep}{3.5mm}{
			\begin{tabular}{c| c| c| c| c}
				\hline
				{Method}&{Handwritten}  & Caltech-7 & Caltech-20 & Citeseer \\
				\hline
		  	    AMGL     & 67.58s &20.12s  &77.63s   &449.07s\\
		  	    MLRSSC   & 52.44s &22.26s  &607.28s  &106.10s\\
		  	    MSC\_IAS &80.78s  &57.18s  &93.87s   &191.29s \\
		  	    LMVSC    &10.55s  &135.79s &342.97s  &21.33s \\		 
		  	    SMVSC    &8.58s   &236.32s  &447.58s &21.82s \\
				\hline				
				SMC     &10.37s  &63.29s  &135.99s  &5.78s
				  \\
				\hline
			\end{tabular}
		}
\end{table}

\begin{table}[htbp] \small
		\caption{Time comparison on graph datasets.}
		\label{Time}
		\centering
		\setlength{\tabcolsep}{7mm}{
			\begin{tabular}{c| c| c| c}
				\hline
				{Method}&{ACM}  & DBLP & IMDB \\
				\hline
		  	    LINE&180.31s&573.54s&486.75s \\
		  	    GAE&286.57s&2672.62s&1886.22s\\
		  	    MNE&94.25s&253.1s&221.53s \\
		  	    PMNE(n)&130.42s&290.84s&365.24s \\		  
		  	    PwMC &174.55s &781.92s& 1453.06s\\
		  	    SwMC &30.06s &195.28s &3300.24s\\
		  	    O2MAC&423.5s &4725.36s &4126.37s \\
		  	    HAN &253.24s &376.27s &289.72s \\
				MvAGC&5.8s&5.19s& 10.38s \\
				\hline				SMC&4.86s&4.92s& 4.54s \\
				\hline
			\end{tabular}
		}
\end{table}

\subsection{Parameter Analysis}
\begin{figure}[!htb]
\includegraphics[width=0.3\linewidth]{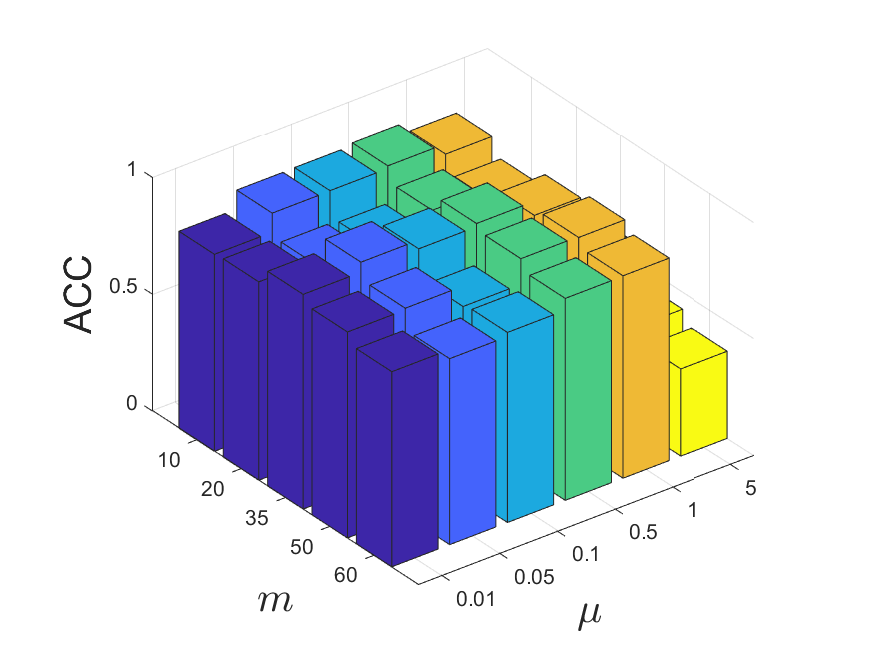}
\includegraphics[width=0.3\linewidth]{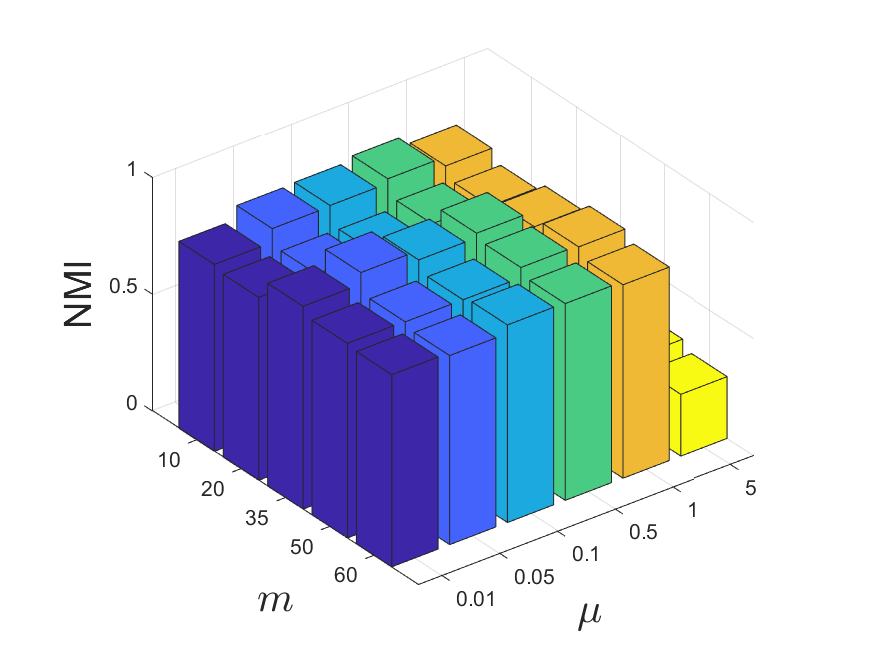}
\includegraphics[width=0.3\linewidth]{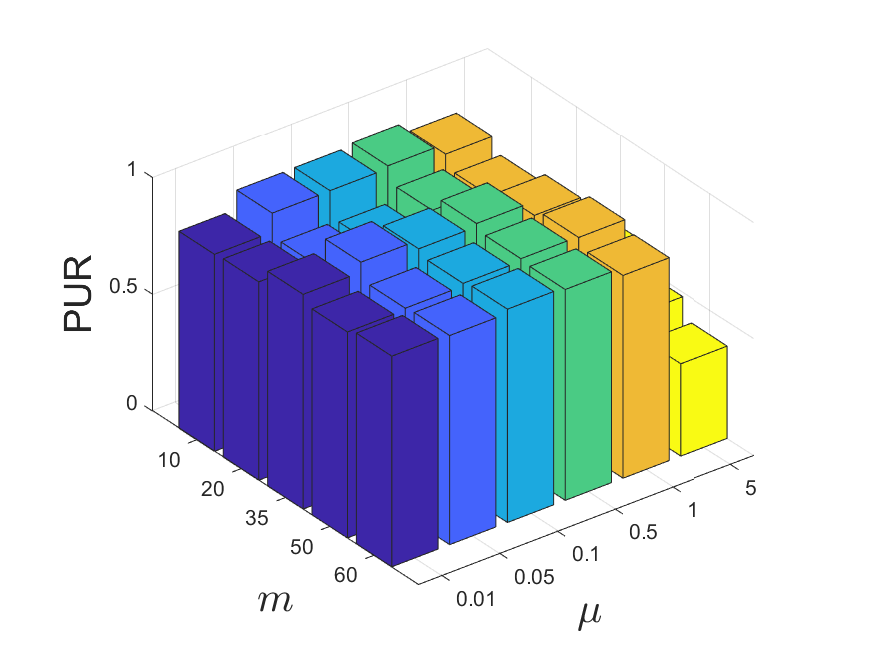}\\
\includegraphics[width=0.3\linewidth]{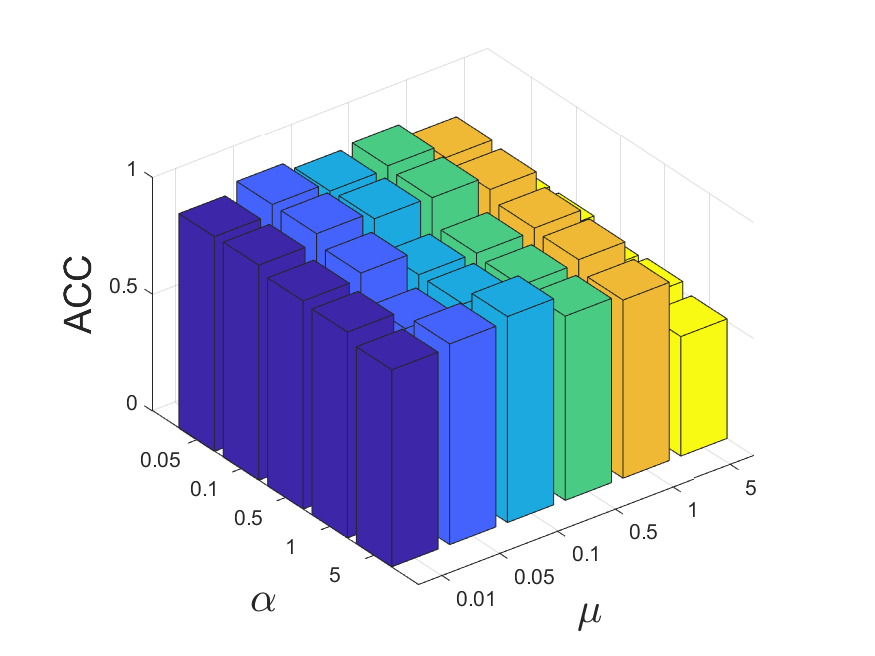}
\includegraphics[width=0.3\linewidth]{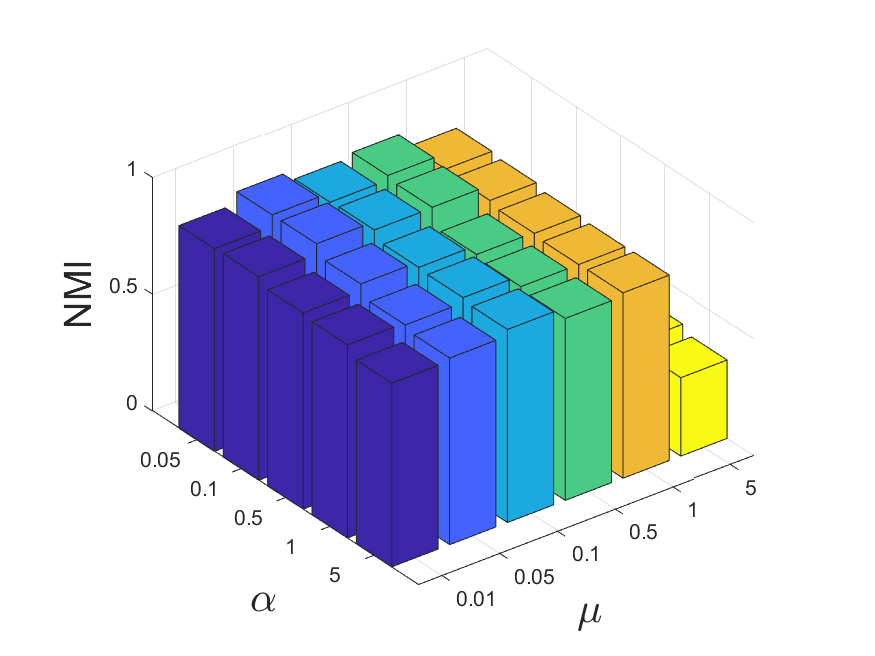}
\includegraphics[width=0.3\linewidth]{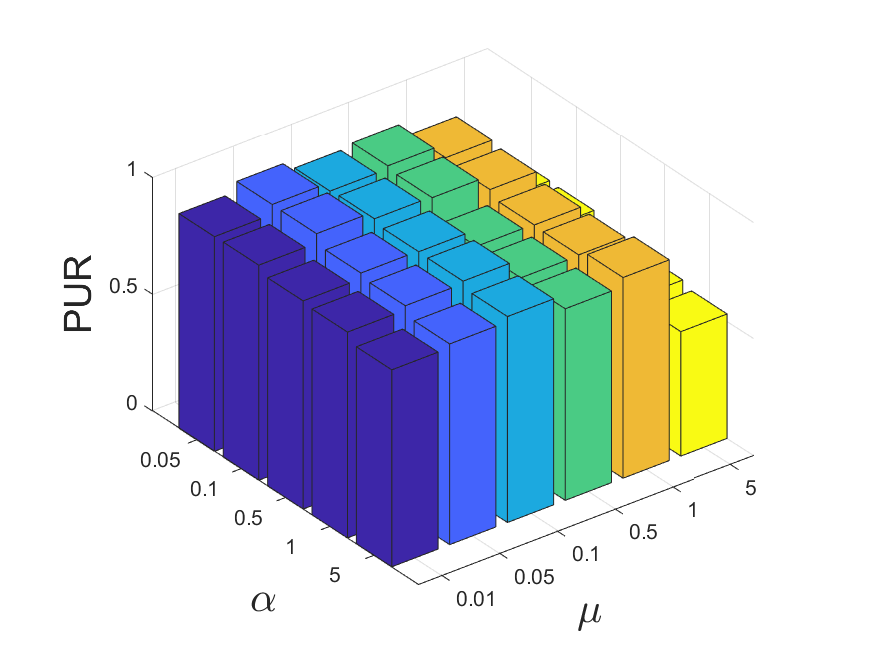}
\caption{The parameter sensitivity on Handwritten.}
	\label{hw-parameter-analysis}
\end{figure}

There are several parameters to tune, including trade-off parameter $\alpha$, filter order $k$, filter parameter $\mu$, and number of anchors $m$. In the multi-view feature data experiment, we set $k=1$ and tune others parameters. Taking Handwritten for example, we show the parameter sensitivity in Fig. \ref{hw-parameter-analysis}. We observe that clustering performance is affected by the anchor number $m$. This makes sense since too many anchors will introduce some noise while too few anchors will fail to represent the whole data. Thus improper $m$ will result in performance degradation. In addition, parameter $\mu$ works well in the range [0.05,0.1,0.5]. Our method also works well for a large range of $\alpha$. 

\begin{figure}[!htb]
\includegraphics[width=0.24\linewidth]{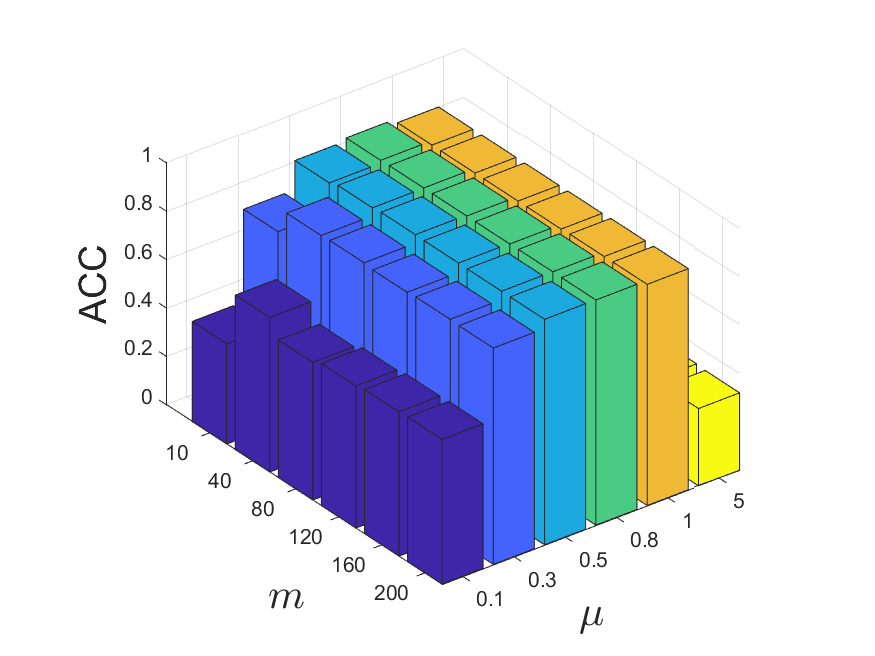}
\includegraphics[width=0.24\linewidth]{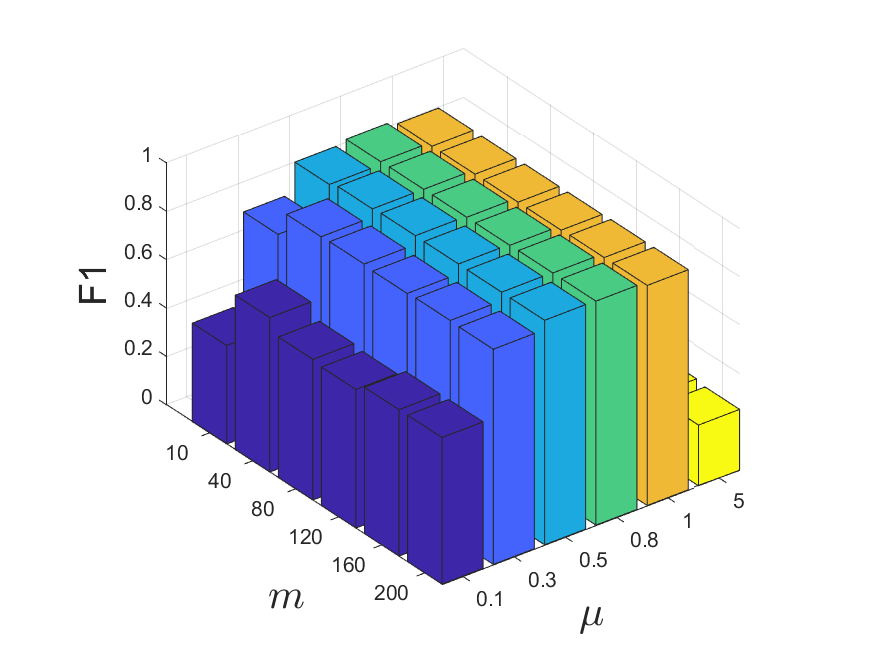}
\includegraphics[width=0.24\linewidth]{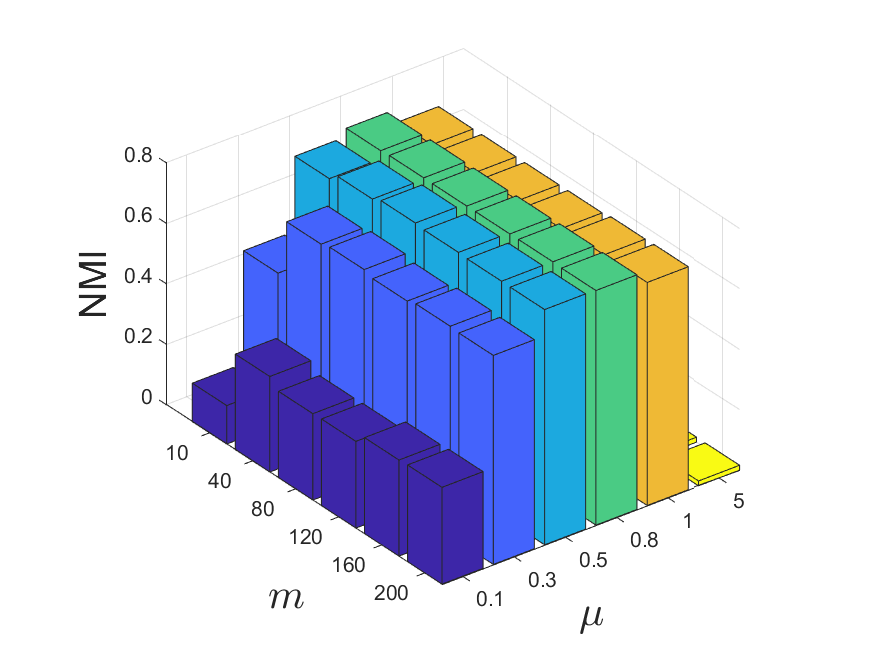}
\includegraphics[width=0.24\linewidth]{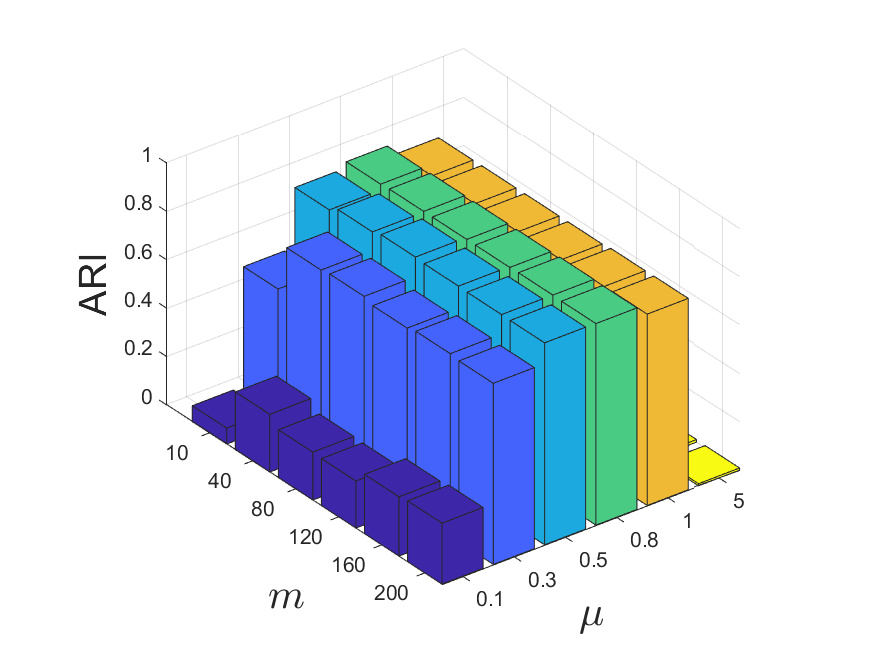}
\\
\includegraphics[width=0.24\linewidth]{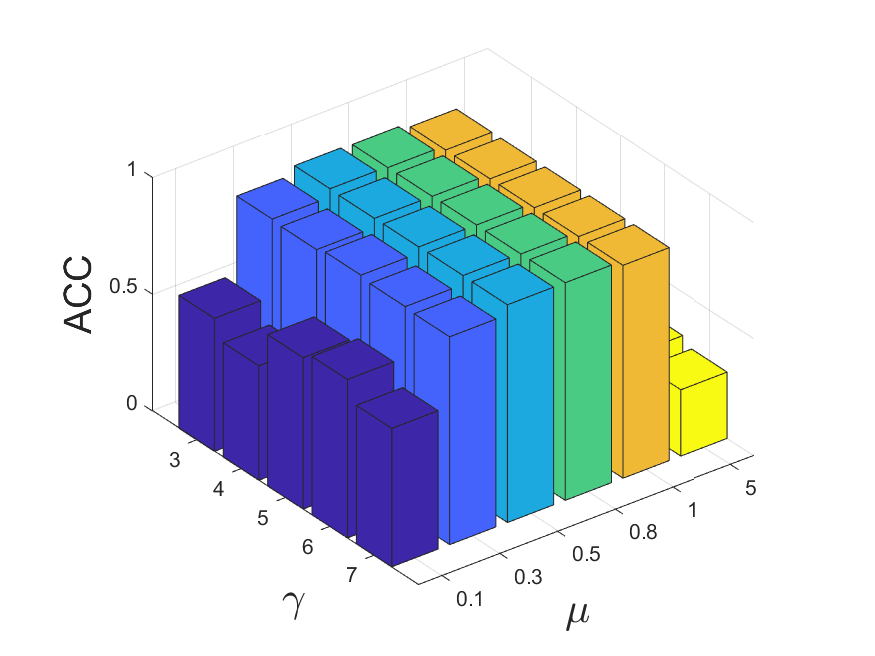}
\includegraphics[width=0.24\linewidth]{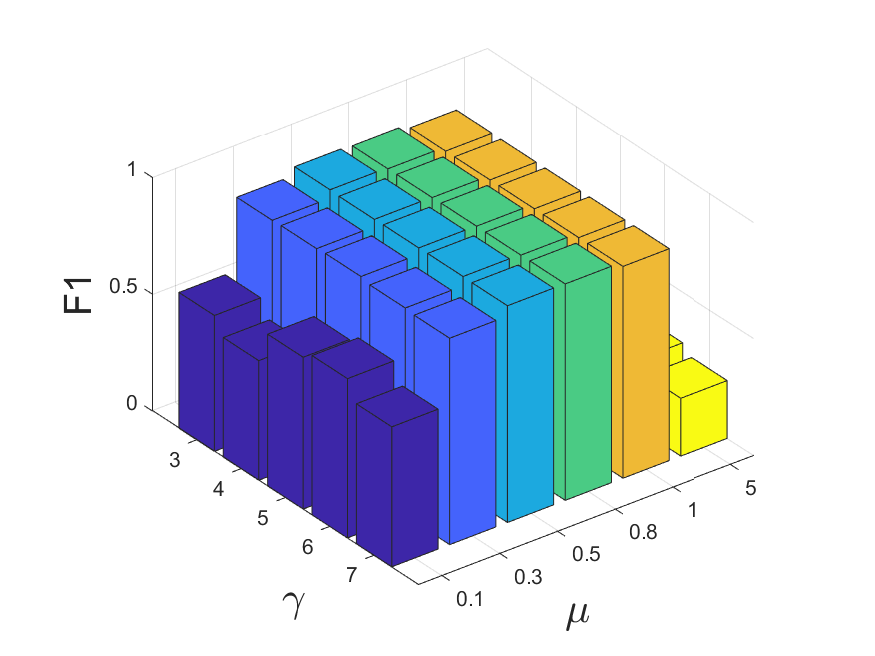}
\includegraphics[width=0.24\linewidth]{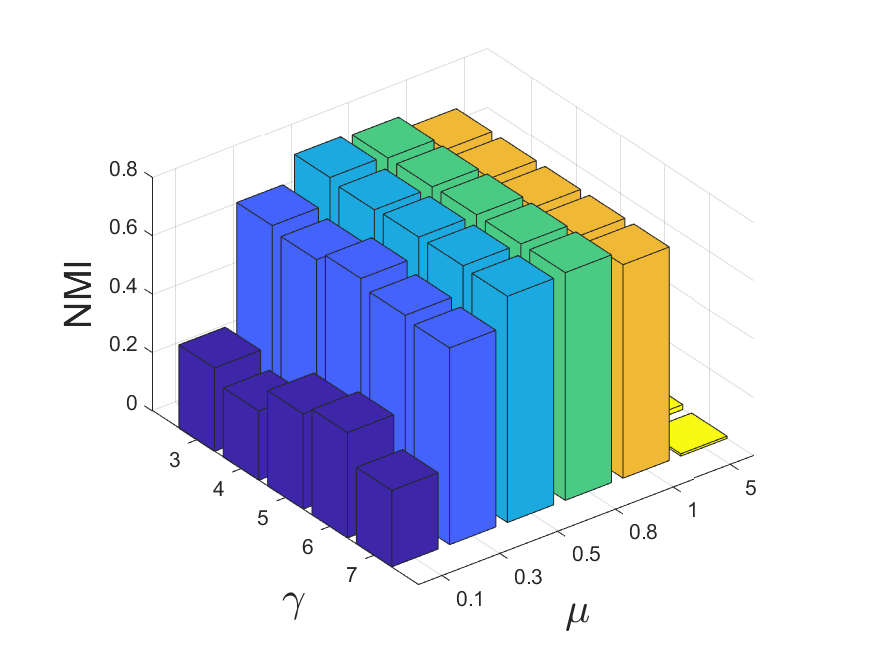}
\includegraphics[width=0.24\linewidth]{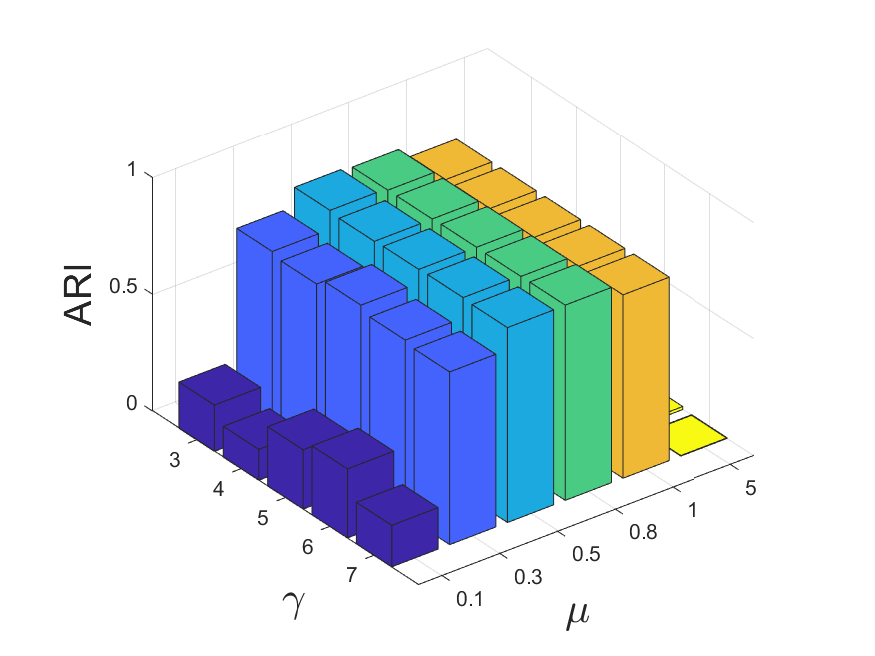}
\\
\includegraphics[width=0.24\linewidth]{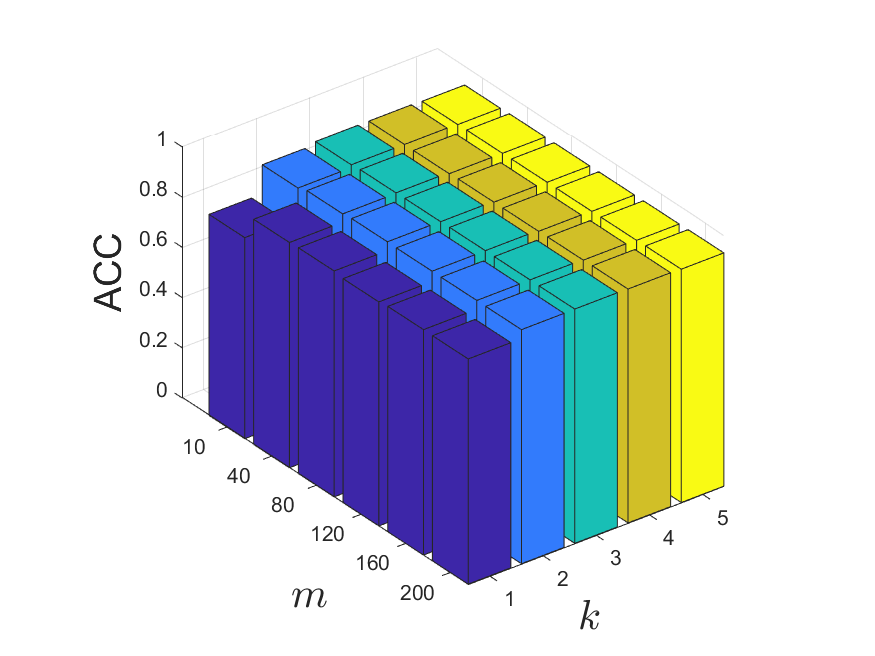}
\includegraphics[width=0.24\linewidth]{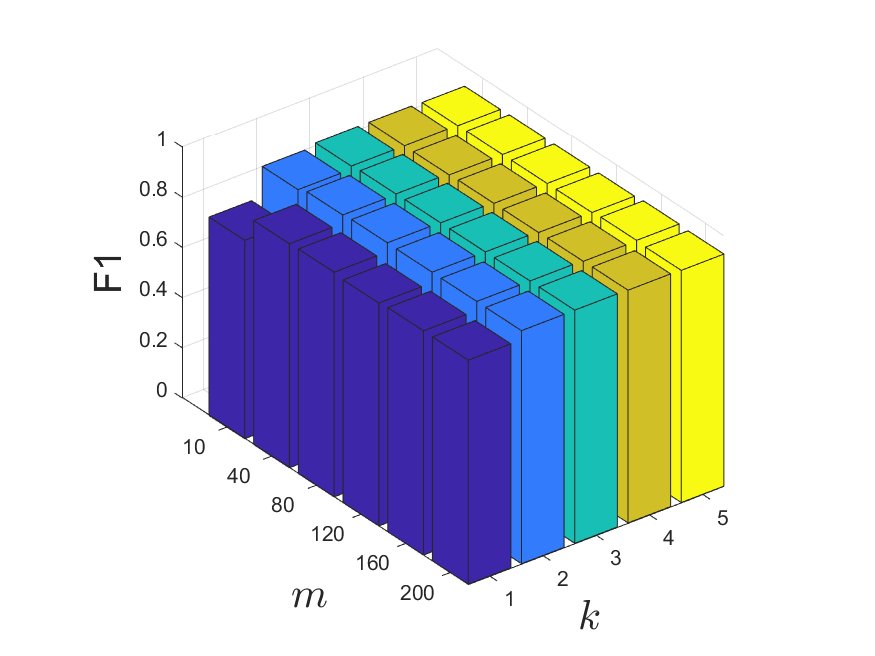}
\includegraphics[width=0.24\linewidth]{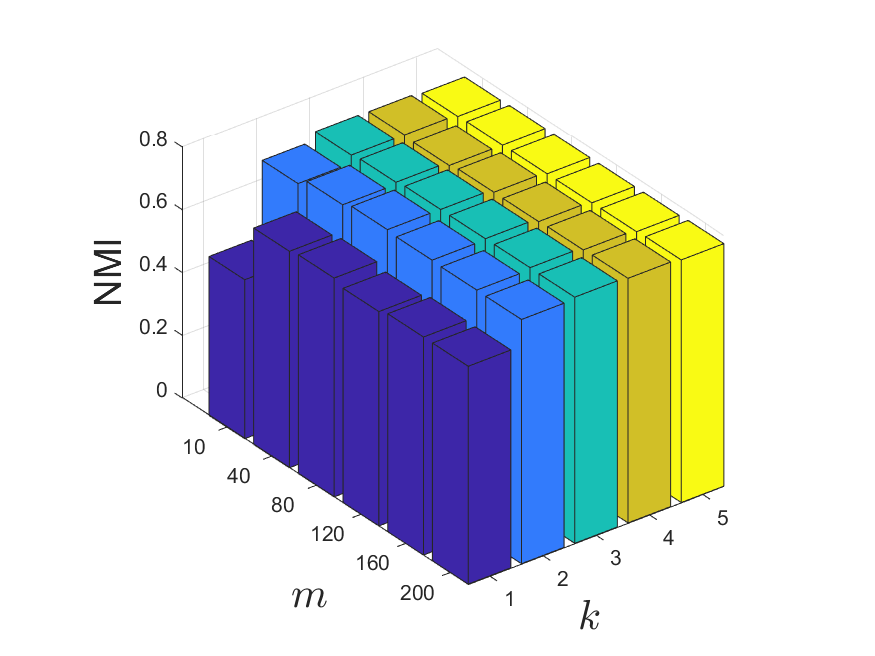}
\includegraphics[width=0.24\linewidth]{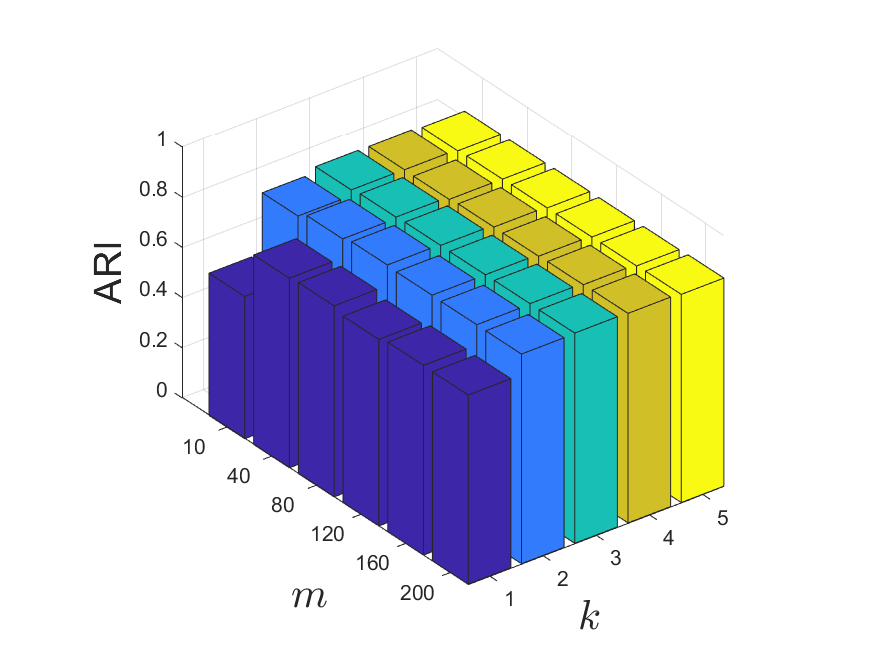}
\caption{The parameter sensitivity on DBLP.}
	\label{dblp-parameter-analysis}
\end{figure}

For heterogeneous graph data, we find that $\alpha$ has little influence on the results, thus we fix it to $20$. DBLP, for instance, we find that its performance is robust to the number $m$ according to Fig. \ref{dblp-parameter-analysis}.
Therefore, we search $m$ in the range of 80 to 120 on all datasets. Besides, it is obvious that the performance is not sensitive to sampling parameter $\gamma$, thus we fix it to 2. Regarding filter order, $k=3$ or $k=4$ often generates good performance. If $k$ is too large, the resulted representation will become too smooth, which makes data points difficult to distinguish. In addition, we can see that either large or small $\mu$ are not good. Since different data have different levels of noise, the value of $\mu$ heavily depends on the specific dataset. 



\begin{figure}[!htbp] \footnotesize
\begin{center}
\begin{tabular}{ccc}
\includegraphics[width=0.25\linewidth]{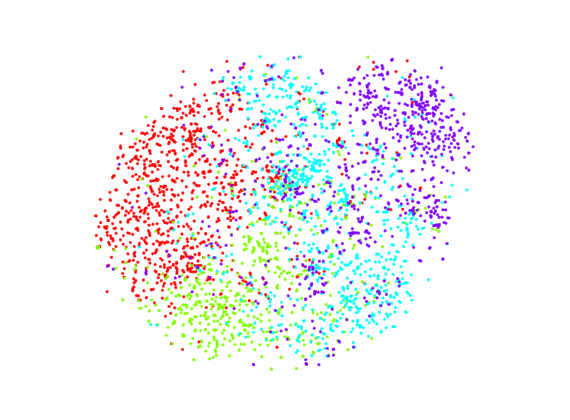} &
\includegraphics[width=0.25\linewidth]{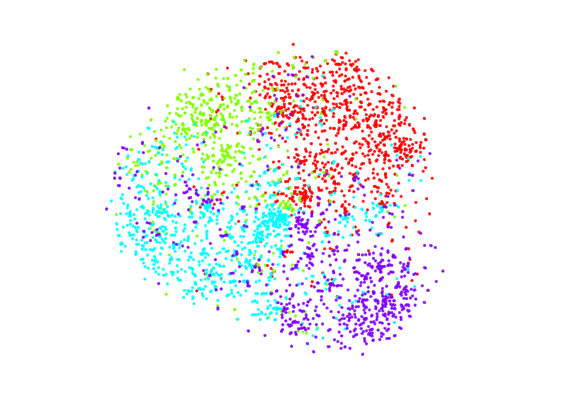} &
\includegraphics[width=0.25\linewidth]{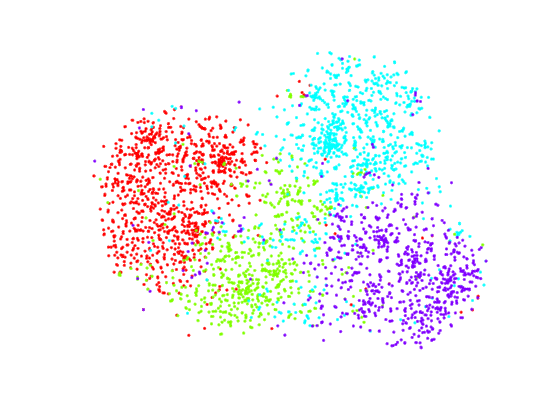} \\
raw feature & $\mu=0.3$ & $\mu=0.5$\\
\\
\includegraphics[width=0.25\linewidth]{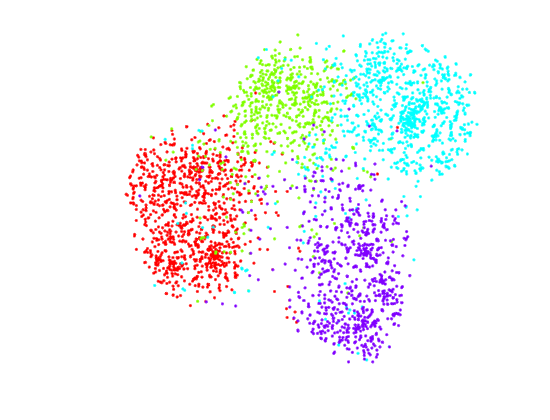} &
\includegraphics[width=0.25\linewidth]{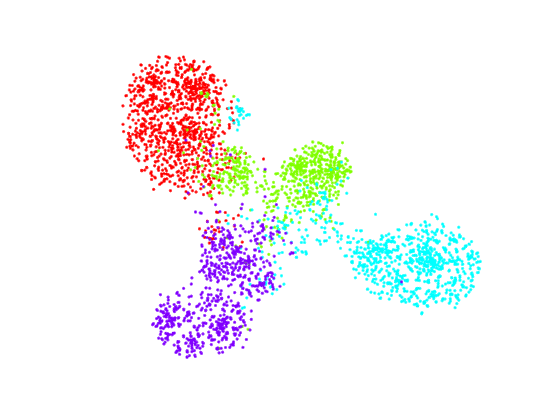} &
\includegraphics[width=0.25\linewidth]{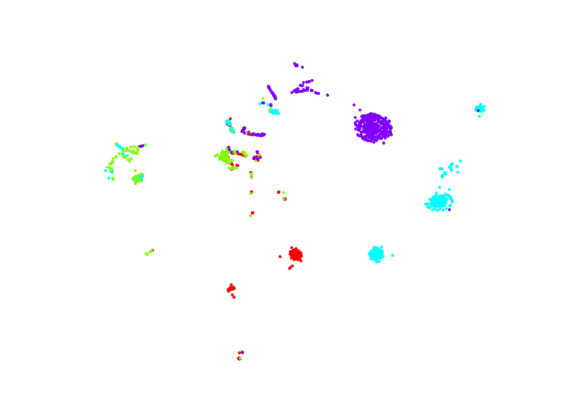} \\
$\mu=0.6$ & $\mu=0.8$ & $\mu=1.0$\\
\end{tabular}
\end{center}
\caption{The 2D visualization of DBLP using t-SNE.}
	\label{Tsne}
\end{figure}

To intuitively see the effect of $\mu$, we fix $k=1$ and display the 2D embedding of DBLP using $t$-SNE algorithm \cite{van2008visualizing}. As shown in Fig. \ref{Tsne}, a proper $\mu$ could produce a clustering-friendly representation.
For example, points from different categories are well separated when $\mu=0.6$ or 0.8. When $\mu=1$, the cluster structure becomes vague.

\section{Conclusion}
In this paper, we propose a scalable graph filter-based multi-view clustering method. It is general to handle both feature and graph data. There are two novel components in the proposed framework. First, an adaptive graph filter is introduced to remove high-frequency noise. Second, a novel sampling mechanism is designed to improve the quality of anchors. Comprehensive experiments demonstrate that the proposed method is not only effective but also efficient. In particular, our proposed method surpasses several state-of-the-art deep neural networks based methods. 

\begin{acknowledgements}
This paper was in part supported by the National Defense Basic Scientific Research Program of China under Grant JCKY2020903B002 and the Natural Science Foundation of China under Grant 61806045.
\end{acknowledgements}


\section*{Compliance with ethical standards}
 \textbf{Conflict of interest} In compliance with ethical standards as researchers, the authors have no potential conflict of interest. The authors certify that they have no affiliations with or involvement in any organization or entity with any financial interest or non-financial interest in the subject matter or materials discussed in this manuscript.

\bibliographystyle{spmpsci}      
\bibliography{template}   

\begin{thebibliography}{10}
\providecommand{\url}[1]{{#1}}
\providecommand{\urlprefix}{URL }
\expandafter\ifx\csname urlstyle\endcsname\relax
  \providecommand{\doi}[1]{DOI~\discretionary{}{}{}#1}\else
  \providecommand{\doi}{DOI~\discretionary{}{}{}\begingroup
  \urlstyle{rm}\Url}\fi

\bibitem{kang2021structured}
Kang, Z., Lin, Z., Zhu, X., Xu, W.: Structured graph learning for scalable
  subspace clustering: From single-view to multi-view.
\newblock IEEE Transactions on Cybernetics  (2021)

\bibitem{chao2021survey}
Chao, G., Sun, S., Bi, J.: A survey on multi-view clustering.
\newblock IEEE Transactions on Artificial Intelligence  (2021)

\bibitem{hou2018safe}
Hou, C., Zeng, L.L., Hu, D.: Safe classification with augmented features.
\newblock IEEE transactions on pattern analysis and machine intelligence
  \textbf{41}(9), 2176--2192 (2018)

\bibitem{yu2021learning}
Yu, Q., Xu, W., Wu, Y., Zhang, H.: Learning to collocate fashion items from
  heterogeneous network using structural and textual features.
\newblock In: International Conference on Neural Computing for Advanced
  Applications, pp. 166--180. Springer (2021)

\bibitem{huang2021robust}
Huang, S., Kang, Z., Xu, Z., Liu, Q.: Robust deep k-means: An effective and
  simple method for data clustering.
\newblock Pattern Recognition \textbf{117}, 107996 (2021)

\bibitem{zhu2019PR}
Zhu, X., Zhu, Y., Zheng, W.: Spectral rotation for deep one-step clustering.
\newblock Pattern Recognition \textbf{105}, 107175 (2020)

\bibitem{mi2022multi}
Mi, Y., Ren, Z., Xu, Z., Li, H., Sun, Q., Chen, H., Dai, J.: Multi-view
  clustering with dual tensors.
\newblock Neural Computing and Applications pp. 1--12 (2022)

\bibitem{liu2021refined}
Liu, Y.: Refined learning bounds for kernel and approximate $ k $-means.
\newblock Advances in Neural Information Processing Systems \textbf{34} (2021)

\bibitem{chen2020multi}
Chen, M.S., Huang, L., Wang, C.D., Huang, D.: Multi-view clustering in latent
  embedding space.
\newblock In: Proceedings of the AAAI conference on artificial intelligence,
  vol.~34, pp. 3513--3520 (2020)

\bibitem{kumar2011cotraining}
Kumar, A., Daum{\'e}, H.: A co-training approach for multi-view spectral
  clustering.
\newblock In: Proceedings of the 28th international conference on machine
  learning (ICML-11), pp. 393--400 (2011)

\bibitem{kumar2011co}
Kumar, A., Rai, P., Daume, H.: Co-regularized multi-view spectral clustering.
\newblock Advances in neural information processing systems \textbf{24},
  1413--1421 (2011)

\bibitem{wang2020parallel}
Wang, H., Yang, Y., Zhang, X., Peng, B.: Parallel multi-view concept clustering
  in distributed computing.
\newblock Neural Computing and Applications \textbf{32}(10), 5621--5631 (2020)

\bibitem{liu2021incomplete}
Liu, X.: Incomplete multiple kernel alignment maximization for clustering.
\newblock IEEE Transactions on Pattern Analysis and Machine Intelligence
  (2021)

\bibitem{kang2020partition}
Kang, Z., Zhao, X., Shi, Peng, C., Zhu, H., Zhou, J.T., Peng, X., Chen, W., Xu,
  Z.: Partition level multiview subspace clustering.
\newblock Neural Networks \textbf{122}, 279--288 (2020)

\bibitem{zhang2017latent}
Zhang, C., Hu, Q., Fu, H., Zhu, P., Cao, X.: Latent multi-view subspace
  clustering.
\newblock In: Proceedings of the IEEE conference on computer vision and pattern
  recognition, pp. 4279--4287 (2017)

\bibitem{li2019deep}
Li, Z., Wang, Q., Tao, Z., Gao, Q., Yang, Z.: Deep adversarial multi-view
  clustering network.
\newblock In: IJCAI, pp. 2952--2958 (2019)

\bibitem{fan2020one2multi}
Fan, S., Wang, X., Shi, C., Lu, E., Lin, K., Wang, B.: One2multi graph
  autoencoder for multi-view graph clustering.
\newblock In: Proceedings of The Web Conference 2020, pp. 3070--3076 (2020)

\bibitem{cheng2020multi}
Cheng, J., Wang, Q., Tao, Z., Xie, D.Y., Gao, Q.: Multi-view attribute graph
  convolution networks for clustering.
\newblock In: IJCAI, pp. 2973--2979 (2020)

\bibitem{Nie2016Parameter}
Nie, F., Li, J., Li, X.: Parameter-free auto-weighted multiple graph learning:
  A framework for multiview clustering and semi-supervised classification.
\newblock In: International Joint Conference on Artificial Intelligence, pp.
  1881--1887 (2016)

\bibitem{Brbi2017Multi}
Brbić, M., Kopriva, I.: Multi-view low-rank sparse subspace clustering.
\newblock Pattern Recognition \textbf{73}, 247--258 (2018)

\bibitem{xia2014robust}
Xia, R., Pan, Y., Du, L., Yin, J.: Robust multi-view spectral clustering via
  low-rank and sparse decomposition.
\newblock In: Proceedings of the AAAI conference on artificial intelligence,
  vol.~28 (2014)

\bibitem{nie2017self}
Nie, F., Li, J., Li, X., et~al.: Self-weighted multiview clustering with
  multiple graphs.
\newblock In: IJCAI, pp. 2564--2570 (2017)

\bibitem{wang2019multi}
Wang, X., Lei, Z., Guo, X., Zhang, C., Shi, H., Li, S.Z.: Multi-view subspace
  clustering with intactness-aware similarity.
\newblock Pattern Recognition \textbf{88}, 50--63 (2019)

\bibitem{kang2019large}
Kang, Z., Zhou, W., Zhao, Z., Shao, J., Han, M., Xu, Z.: Large-scale multi-view
  subspace clustering in linear time.
\newblock In: Proceedings of the AAAI Conference on Artificial Intelligence,
  vol.~34, pp. 4412--4419 (2020)

\bibitem{chen2021smoothed}
Chen, P., Liu, L., Ma, Z., Kang, Z.: Smoothed multi-view subspace clustering.
\newblock In: International Conference on Neural Computing for Advanced
  Applications, pp. 128--140. Springer (2021)

\bibitem{liu2017principled}
Liu, W., Chen, P.Y., Yeung, S., Suzumura, T., Chen, L.: Principled multilayer
  network embedding.
\newblock In: 2017 IEEE International Conference on Data Mining Workshops
  (ICDMW), pp. 134--141. IEEE (2017)

\bibitem{zhang2018scalable}
Zhang, H., Qiu, L., Yi, L., Song, Y.: Scalable multiplex network embedding.
\newblock In: IJCAI, vol.~18, pp. 3082--3088 (2018)

\bibitem{HAN}
Wang, X., Ji, H., Shi, C., Wang, B., Ye, Y., Cui, P., Yu, P.S.: Heterogeneous
  graph attention network.
\newblock In: The World Wide Web Conference, pp. 2022--2032 (2019)

\bibitem{lin2021graph}
Lin, Z., Kang, Z.: Graph filter-based multi-view attributed graph clustering.
\newblock In: Proceedings of the Thirtieth International Joint Conference on
  Artificial Intelligence, {IJCAI-21}, pp. 2723--2729 (2021)

\bibitem{lin2021multi}
Lin, Z., Kang, Z., Zhang, L., Tian, L.: Multi-view attributed graph clustering.
\newblock IEEE Transactions on Knowledge and Data Engineering  (2021)

\bibitem{dong2019learning}
Dong, X., Thanou, D., Rabbat, M., Frossard, P.: Learning graphs from data: A
  signal representation perspective.
\newblock IEEE Signal Processing Magazine \textbf{36}(3), 44--63 (2019)

\bibitem{pan2021multi}
Pan, E., Kang, Z.: Multi-view contrastive graph clustering.
\newblock Advances in Neural Information Processing Systems \textbf{34} (2021)

\bibitem{ma2020towards}
Ma, Z., Kang, Z., Luo, G., Tian, L., Chen, W.: Towards clustering-friendly
  representations: Subspace clustering via graph filtering.
\newblock In: Proceedings of the 28th ACM International Conference on
  Multimedia, pp. 3081--3089 (2020)

\bibitem{lv2021pseudo}
Lv, J., Kang, Z., Lu, X., Xu, Z.: Pseudo-supervised deep subspace clustering.
\newblock IEEE Transactions on Image Processing \textbf{30}, 5252--5263 (2021)

\bibitem{zhang2020twin}
Zhang, Z., Sun, Y., Wang, Y., Zhang, Z., Zhang, H., Liu, G., Wang, M.:
  Twin-incoherent self-expressive locality-adaptive latent dictionary pair
  learning for classification.
\newblock IEEE Transactions on Neural Networks and Learning Systems
  \textbf{32}(3), 947--961 (2020)

\bibitem{nie2016constrained}
Nie, F., Wang, X., Jordan, M., Huang, H.: The constrained laplacian rank
  algorithm for graph-based clustering.
\newblock In: Proceedings of the AAAI conference on artificial intelligence,
  vol.~30 (2016)

\bibitem{tang2015line}
Tang, J., Qu, M., Wang, M., Zhang, M., Yan, J., Mei, Q.: Line: Large-scale
  information network embedding.
\newblock In: Proceedings of the 24th international conference on world wide
  web, pp. 1067--1077 (2015)

\bibitem{Kipf2016VariationalGA}
Kipf, T.N., Welling, M.: Variational graph auto-encoders.
\newblock In: NIPS Workshop on Bayesian Deep Learning (2016)

\bibitem{van2008visualizing}
Van~der Maaten, L., Hinton, G.: Visualizing data using t-sne.
\newblock Journal of machine learning research \textbf{9}(11) (2008)

\end{thebibliography}

%
%

\end{document}